\documentclass[runningheads]{llncs}

\usepackage{lmodern}

\usepackage{accv}

\usepackage{accvabbrv}

\usepackage{graphicx}
\usepackage{booktabs}

\usepackage[accsupp]{axessibility}  

\usepackage{hyperref}

\usepackage{orcidlink}

\usepackage{amsmath,amssymb,bm}
\usepackage{tikz}
\usetikzlibrary{arrows.meta,positioning,shapes.geometric,calc,fit,backgrounds,decorations.pathreplacing,patterns}

\begin{document}

\title{CLIFT: Contrastive LiDAR Instance Features for 3D Pedestrian Tracking} 

\titlerunning{CLIFT: Contrastive LiDAR Instance Features for 3D Pedestrian Tracking}

\author{German Devchich\orcidlink{0009-0003-8000-4342} \and
Timur Akhtyamov\orcidlink{0000-0003-0877-3318} \and
Denis Fatykhoph\orcidlink{0009-0004-3998-1915} \and
Gonzalo Ferrer\orcidlink{0000-0003-2704-7186}}

\authorrunning{G.~Devchich et al.}

\institute{Applied AI Institute, Moscow, Russia\\
\email{\{g.devchich,t.akhtyamov,d.fatykhoph,g.ferrer\}@applied-ai.ru}}

\maketitle

\begin{abstract}
A social robot has to keep track of the people around it over time to plan its motion, a task handled by 3D multi-object tracking. For the LiDAR modality, trackers usually sit on top of a detector and link its detections across frames with a separate motion model or re-identification network. We instead show how a 3D LiDAR detector itself can be turned into a tracker by adding an instance-level contrastive objective on its intermediate voxel features. Each embedding is read from the voxels inside a detected box, and boxes are matched across frames by cosine similarity and the Hungarian algorithm, without a motion model. The detector backbone is fine-tuned together with this objective, which learns identity-aware features while leaving detection accuracy essentially intact. Because the method uses only the voxel features that LiDAR detectors already produce, it is backbone-agnostic. On the JRDB 3D tracking leaderboard our LiDAR-only tracker ranks first on HOTA (38.58) and IDF1 (37.99), 13.04 and 9.30 points above the next published method, and second on MOTA and OSPA. Code is available at \url{https://hermak56.github.io/CLIFT/}.
\end{abstract}

\section{Introduction} \label{sec:introduction}


Safe motion planning for a mobile robot in a crowd depends on stable per-person trajectories over time. Producing them is the task of 3D multi-object tracking (MOT), which runs on top of a detector and links its per-frame boxes into identity-consistent tracks. On LiDAR this is harder than in images: point clouds are sparse, and a pedestrian is a small, deformable object whose points change from frame to frame even when the person barely moves. LiDAR-based 3D MOT therefore lags behind 2D tracking on benchmarks such as JRDB~\cite{jrdb}. Detection accuracy has improved~\cite{dccla, rpea}, but tracking has not followed: past a certain detection quality most errors are identity switches and broken tracks, so the limit is how stable the per-object features are over time, not localization.

Existing 3D MOT either learns no identity features, as in Kalman-plus-Hungarian pipelines~\cite{ab3dmot}, or learns them from raw point clouds or cropped objects~\cite{pcdan, ss3dmot}, where sparsity and occlusion hit the input directly and JRDB results stay limited. A common alternative fuses a camera with the LiDAR~\cite{jrmot, mmpat}. This raises identity metrics, but it comes at a cost: the two sensors need accurate extrinsic calibration and temporal synchronization, the camera branch adds parameters and inference latency, and the system inherits the failure modes of the image modality under poor lighting. Fusion is a standard design choice here rather than one shown to be necessary once LiDAR alone is pushed. Contrastive learning solves a similar association problem in 2D~\cite{contrastr}, but it is unclear whether it carries over to LiDAR, whose features change with the same viewpoint and sparsity effects that break identity.

We apply contrastive learning to the intermediate voxel features of a trained LiDAR detector, which are already shaped for semantic consistency, instead of to raw geometry. Each embedding is pooled from the voxels inside a predicted box, trained with InfoNCE~\cite{infonce}, and matched across frames by cosine similarity and the Hungarian algorithm, with no motion model. The backbone is used frozen or fine-tuned jointly with the tracker, which leaves detection essentially intact. On the JRDB leaderboard this LiDAR-only tracker ranks first on the association-oriented HOTA and IDF1 metrics, above every published method, with the gains concentrated on identity.

\section{Related Work} \label{sec:related_work}

\subsection{3D Detection and Tracking}

Multi-object tracking is dominated by tracking-by-detection, where per-frame detections are linked across time by motion and association rules. SORT~\cite{sort} paired a Kalman filter with the Hungarian algorithm, and DeepSORT~\cite{deepsort} added a learned appearance embedding that reduced identity switches under occlusion. ByteTrack~\cite{bytetrack} later improved crowded-scene tracking by recovering low-confidence detections in a second association pass. A second line folds detection and association into one network: Tracktor~\cite{tracktor} reuses a detector's box regressor for propagation, JDE~\cite{jde} and FairMOT~\cite{fairmot} learn an identity embedding jointly with detection in a shared backbone, and transformer trackers such as TrackFormer~\cite{trackformer} and CenterTrack~\cite{centertrack} attach association to the detector directly. These joint models reach strong 2D results but need large tracking-annotated datasets, and the embedding branch has to be balanced against detection during training, a tension FairMOT~\cite{fairmot} analyzes in detail.

In 3D, classical tracking-by-detection remains the default. AB3DMOT~\cite{ab3dmot} combines a 3D Kalman filter with Hungarian assignment and learns no identity features, CenterPoint~\cite{centerpoint} predicts centers and inter-frame offsets, and ImmortalTracker~\cite{immortaltracker} shows that track-lifecycle handling can matter more than association logic. A smaller set of methods learns the association step: GNN3DMOT~\cite{gnn3dmot} propagates object features through a graph neural network so that detections inform each other, and 3DMOTFormer~\cite{3dmotformer} casts association as edge classification on a track-detection graph transformer. Both raise association quality on driving benchmarks but keep a hand-designed motion or geometry prior at their core. Sparse-convolutional LiDAR detectors, from Person-MinkUNet~\cite{personminkunet} to the DCCLA and RPEA backbones used in this work~\cite{dccla, rpea}, produce accurate boxes and, more useful for this work, intermediate voxel features already trained to separate pedestrians from their surroundings.

\subsection{Pedestrian Tracking for Social Robotics}

Social robot navigation depends on persistent pedestrian identities, since identity switches and fragmented tracks propagate into pedestrian motion prediction and planning~\cite{akhtyamov2023social}. Robot-mounted pedestrian LiDAR is sparser and more frequently occluded than the vehicle-mounted data of autonomous-driving benchmarks, and it is underrepresented by comparison. JRDB~\cite{jrdb} is the reference benchmark for this setting, with egocentric mobile-robot data, 360-degree LiDAR, time-consistent identity labels, and a public leaderboard whose metrics separate detection from association (HOTA~\cite{hota}, MOTA~\cite{clearmot}, IDF1~\cite{idf1}, OSPA~\cite{ospa}). The strongest published JRDB 3D trackers are classical tracking-by-detection pipelines: ELPTNet~\cite{elptnet}, built on the same LiDAR detector this work uses, engineers a motion model and an association cost from geometric, appearance, and motion cues, and S2MAT~\cite{s2mat} couples tracking with static mapping. Learning-based LiDAR trackers such as PC-DAN~\cite{pcdan} and SS3D-MOT~\cite{ss3dmot} learn association features, but they operate on raw point clouds or object-level crops directly exposed to sparsity and occlusion, and their JRDB results stay limited. Multimodal fusion~\cite{jrmot, mmpat} raises identity-oriented metrics by adding camera appearance, at the cost of calibration and compute, and is often adopted by default rather than after testing how far single-modality LiDAR can go.

\subsection{Contrastive Identity Learning}

Contrastive objectives learn representations by pulling paired views of the same instance together and pushing different instances apart. InfoNCE~\cite{infonce} provides the objective behind most modern variants, SimCLR~\cite{simclr} and MoCo~\cite{moco} scale it for visual pretraining, and VICReg~\cite{vicreg} reaches similar representations without explicit negatives. The supervised form of the loss~\cite{supcon} admits several positives per anchor when labels are available, which is the setting this work uses once ground-truth identities define the positive set. Applied to tracking, dense contrastive sampling was shown by QDTrack~\cite{qdtrack} to produce an embedding space where nearest-neighbor matching alone is enough for association, and GHOST~\cite{ghost} reported that a well-adapted appearance model with a simple association rule matches heavier end-to-end trackers. ContrasTR~\cite{contrastr} adds an instance-level contrastive loss to a transformer detector and learns identity embeddings from detector features in the 2D case. In 3D, contrastive objectives have mostly served self-supervised pretraining of point and voxel encoders, as in PointContrast~\cite{pointcontrast} and the LiDAR segment discrimination of SegContrast~\cite{segcontrast}, and more recently identity learning for camera-based 3D tracking queries~\cite{syncl}. Whether the idea transfers to LiDAR identity tracking is not settled.

This work transfers contrastive identity learning to 3D by applying contrast to intermediate voxel-level detector features rather than to raw crops, and by jointly fine-tuning the shared backbone under the resulting conflict between detection and tracking. That conflict is the subject of the multi-task optimization literature, where competing task gradients are handled by projection~\cite{pcgrad}, similarity targets~\cite{gradvac}, or constrained descent~\cite{cagrad}, and task losses are balanced by gradient norms~\cite{gradnorm}, homoscedastic uncertainty~\cite{kendall}, or a multi-objective solver~\cite{mgda}. We find that a plain fixed loss weight is enough here, and report that negative result in Section~\ref{sec:joint}. To our knowledge, intermediate detector-level voxel features have not been used as the basis for contrastive identity learning in 3D MOT.

\section{Materials and Methods} \label{sec:method}

This section fixes the notation and builds the proposed method on top of a tracking-by-detection setup. Figure~\ref{fig:pipeline} gives the overall layout.


\definecolor{detblue}{HTML}{003D6E}
\definecolor{trackyellow}{HTML}{F5A623}
\definecolor{idblue}{HTML}{1F77B4}
\definecolor{frozenfill}{HTML}{E8ECF1}
\definecolor{ourfill}{HTML}{FFF4D6}
\definecolor{voxelfill}{HTML}{B8C9DD}
\definecolor{voxelinside}{HTML}{4F7FB0}
\definecolor{neutralfill}{HTML}{F5F5F5}
\definecolor{neutraledge}{HTML}{6A6A6A}
\definecolor{fgdark}{HTML}{4A4A4A}
\definecolor{fglabel}{HTML}{7A7A7A}

\providecommand{\figtitle}[1]{{\sffamily\bfseries\small #1}}
\providecommand{\figtag}[1]{{\sffamily\footnotesize\itshape\color{fglabel}#1}}
\providecommand{\figflow}[1]{{\sffamily\footnotesize\color{fgdark} #1}}

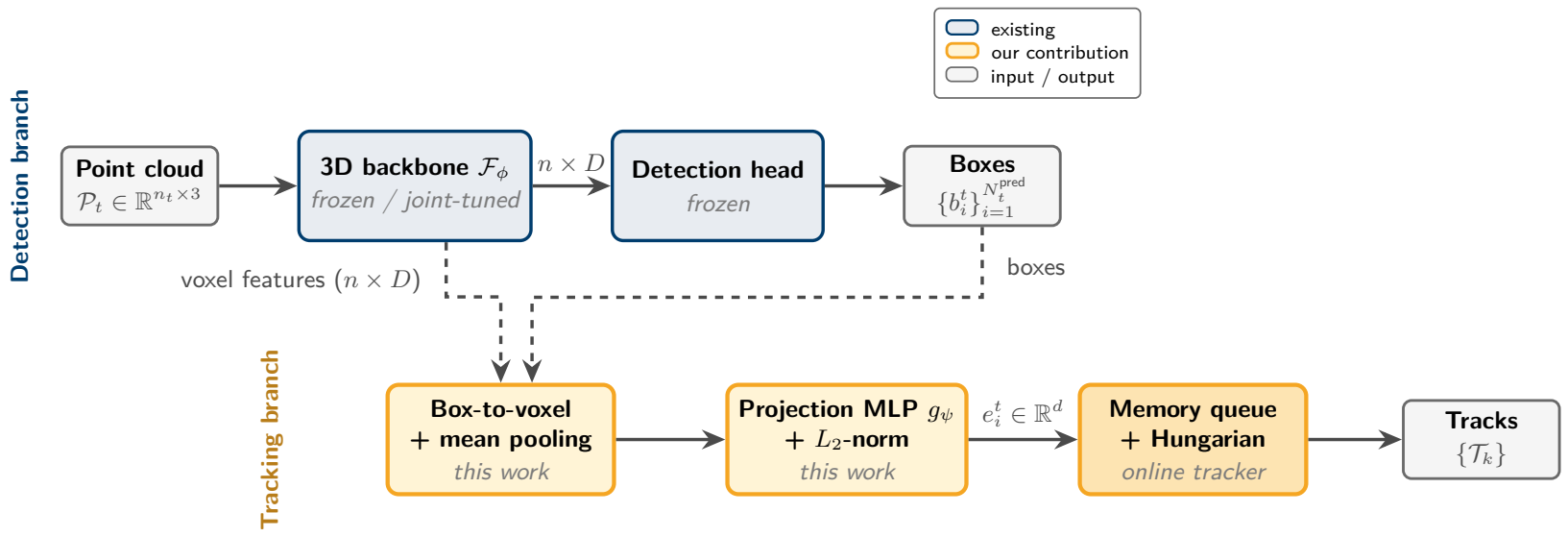
\begin{figure}[t]
\centering
\resizebox{0.88\linewidth}{!}{%
\begin{tikzpicture}[
    font=\sffamily,
    >=Stealth,
    frozenblock/.style={
        draw=detblue, line width=1.2pt, fill=frozenfill, rounded corners=4pt,
        minimum width=2.7cm, minimum height=1.4cm, align=center, inner sep=4pt},
    ourblock/.style={
        draw=trackyellow, line width=1.5pt, fill=ourfill, rounded corners=4pt,
        minimum width=2.9cm, minimum height=1.4cm, align=center, inner sep=4pt},
    iolock/.style={
        draw=neutraledge, line width=1pt, fill=neutralfill, rounded corners=3pt,
        minimum width=2.0cm, minimum height=1.0cm, align=center, inner sep=3pt},
    flow/.style={-{Stealth[length=3mm,width=2.5mm]}, line width=1.2pt, draw=fgdark},
    vconnect/.style={-{Stealth[length=3mm,width=2.5mm]}, line width=1.2pt, draw=fgdark, dashed}
]

\node[iolock] (input) at (0,0)
    {\figtitle{Point cloud}\\[1pt]\figflow{$\mathcal{P}_t \in \mathbb{R}^{n_t \times 3}$}};
\node[frozenblock, right=10mm of input] (backbone)
    {\figtitle{3D backbone $\mathcal{F}_\phi$}\\[1pt]\figtag{frozen / joint-tuned}};
\node[frozenblock, right=10mm of backbone] (dethead)
    {\figtitle{Detection head}\\[1pt]\figtag{frozen}};
\node[iolock, right=10mm of dethead] (bboxes)
    {\figtitle{Boxes}\\[1pt]\figflow{$\{b_i^t\}_{i=1}^{N^{\text{pred}}_t}$}};

\draw[flow] (input.east)    -- (backbone.west);
\draw[flow] (backbone.east) -- node[above,yshift=1pt]{\figflow{$n \times D$}} (dethead.west);
\draw[flow] (dethead.east)  -- (bboxes.west);

\node[ourblock, below=18mm of backbone, xshift=11mm] (boxvoxel)
    {\figtitle{Box-to-voxel}\\\figtitle{+ mean pooling}\\[1pt]\figtag{this work}};
\node[ourblock, right=14mm of boxvoxel] (mlp)
    {\figtitle{Projection MLP $g_\psi$}\\\figtitle{+ $L_2$-norm}\\[1pt]\figtag{this work}};
\node[ourblock, right=14mm of mlp, fill=ourfill!80!trackyellow] (tracker)
    {\figtitle{Memory queue}\\\figtitle{+ Hungarian}\\[1pt]\figtag{online tracker}};
\node[iolock, right=12mm of tracker] (tracks)
    {\figtitle{Tracks}\\[1pt]\figflow{$\{\mathcal{T}_k\}$}};

\draw[flow] (boxvoxel.east) -- (mlp.west);
\draw[flow] (mlp.east) -- node[above,yshift=1pt]{\figflow{$e_i^t \in \mathbb{R}^{d}$}} (tracker.west);
\draw[flow] (tracker.east) -- (tracks.west);

\draw[vconnect] ($(backbone.south)+(4mm,0)$) -- ++(0,-7mm) -| (boxvoxel.north);
\node[font=\sffamily\footnotesize, text=fgdark, anchor=east]
    at ($(backbone.south)+(2mm,-5mm)$) {voxel features ($n \times D$)};
\draw[vconnect] (bboxes.south) -- ++(0,-9mm) -| ($(boxvoxel.north)+(4mm,0)$);
\node[font=\sffamily\footnotesize, text=fgdark, anchor=west]
    at ($(bboxes.south)+(2mm,-5mm)$) {boxes};

\node[anchor=east, font=\sffamily\bfseries\small, text=detblue]
    at ($(input.west)+(-3mm,0)$) {\rotatebox{90}{Detection branch}};
\node[anchor=east, font=\sffamily\bfseries\small, text=trackyellow!75!black]
    at ($(boxvoxel.west)+(-12mm,0)$) {\rotatebox{90}{Tracking branch}};

\begin{scope}[shift={(11.5,1.7)}]
    \node[draw=neutraledge, rounded corners=2pt, line width=0.6pt, fill=white,
          inner sep=4pt, align=left, font=\sffamily\scriptsize] {
        \tikz{\fill[frozenfill, draw=detblue, line width=0.8pt] (0,0) rectangle (0.4,0.18);} \;existing\\[0.5pt]
        \tikz{\fill[ourfill, draw=trackyellow, line width=1pt] (0,0) rectangle (0.4,0.18);} \;our contribution\\[0.5pt]
        \tikz{\fill[neutralfill, draw=neutraledge, line width=0.6pt] (0,0) rectangle (0.4,0.18);} \;input / output
    };
\end{scope}

\end{tikzpicture}%
}
\caption{The CLIFT pipeline. A single forward pass through the 3D detector produces both predicted boxes (top, detection branch) and the voxel-level feature map. For each box, the tracking branch (bottom, this work) gathers the voxels inside it, mean-pools their features, projects and $L_2$-normalizes them into an instance embedding $e_i^t$, and the online tracker associates embeddings across frames by cosine similarity and the Hungarian algorithm. The backbone is either frozen or jointly fine-tuned; the detection head is never modified.}
\label{fig:pipeline}
\end{figure}


\subsection{Problem Formulation} \label{sec:problem}

A LiDAR scan at time $t$ is a point cloud $\mathcal{P}_t = \{p_1, \dots, p_{N_t}\}$ with $p_k \in \mathbb{R}^3$. After voxelization with voxel size $\Delta$, each point is mapped to a voxel index $v \in \mathbb{Z}^3$, and $\mathcal{V}_t$ is the set of non-empty voxels at frame $t$. Each voxel index $v$ has a known spatial center $c(v) = \Delta \odot v + \tfrac{1}{2}\Delta \in \mathbb{R}^3$, where $\odot$ is the element-wise (Hadamard) product.

A 3D bounding box is $b = (x, y, z, l, w, h, \theta)$, with center $(x,y,z)$, size $(l,w,h)$, and heading $\theta$. A LiDAR 3D pedestrian detector with parameters $\phi$, trained solely for the detection task~\cite{dccla, rpea}, produces
\begin{equation}
\mathcal{D}_\phi : \mathcal{P}_t \mapsto \mathcal{B}_t = \{(b_i, s_i)\}_{i=1}^{N_t},
\label{eq:detector}
\end{equation}
and we take its training protocol as given. Predicted boxes carry a confidence $s_i \in [0,1]$ but no identity. Ground-truth annotations, available only at training time, are pairs $(\hat b_i, \hat\tau_i)$ of a box and a track identifier $\hat\tau_i \in \mathbb{N}$ that stays constant across frames for the same physical pedestrian.

A track is a time-ordered sequence of boxes that share one identifier. Writing $b^t_i$ for the box of pedestrian $i$ at frame $t$, the track of pedestrian $i$ is
\begin{equation}
\mathcal{T}_i = \big( b^t_i \big)_{t \in \mathcal{O}_i},
\label{eq:track}
\end{equation}
where $\mathcal{O}_i$ is the set of frames in which $i$ has been observed. We write $\mathcal{R}_t$ for the set of tracks active at frame $t$. Assigning detector boxes to tracks is the tracking problem: at every frame, detections in $\mathcal{B}_t$ are matched against the active tracks $\mathcal{R}_{t-1}$.

On pedestrian-centric LiDAR benchmarks, tracking quality depends on the cross-frame association step far more than on further gains in $\mathcal{D}_\phi$, and association depends in turn on a per-detection representation that is stable across frames for one pedestrian and discriminative between different ones. Constructing that representation is what the rest of the section addresses.

\subsection{Instance Embeddings from Detector Features} \label{sec:embeddings}

Assuming a LiDAR detector is decomposed internally into a voxel-level feature extractor and a detection head, let $\mathcal{F}_\phi$ denote the feature extractor:
\begin{equation}
\mathcal{F}_\phi : \mathcal{V}_t \mapsto \{f_v \in \mathbb{R}^D : v \in \mathcal{V}_t\},
\label{eq:feature_extractor}
\end{equation}
that is, $\mathcal{F}_\phi$ maps each non-empty voxel $v$ to a feature vector $f_v$ of dimension $D$. We read these features at the last voxel-level layer before the detection head, rather than from raw points or cropped boxes: there they are shaped by the full detection pipeline yet still voxel-level, retaining spatial structure while carrying detection-level semantics. This transfers to 3D LiDAR the idea of ContrasTR~\cite{contrastr}, which learns identity embeddings from detector features in 2D.

The boxes that define these voxel sets differ between the two regimes. At training time we pool from the ground-truth boxes, whose track identifiers come directly from the training annotation. No matching between predictions and annotations is performed, so the identity supervision carries no association noise. At inference the annotation is unavailable and the same pooling is applied to the detector's predicted boxes. Writing $b_i$ for the box of instance $i$ at frame $t$ under either regime, the set of voxels falling inside it is
\begin{equation}
\mathcal{V}_t^i = \{v \in \mathcal{V}_t : c(v) \in b_i\},
\end{equation}
using the voxel centers $c(v)$ from Section~\ref{sec:problem}. The box location is already given, so $\mathcal{V}_t^i$ is known without a separate voxel-to-instance assignment step.

Boxes are dilated before pooling to recover voxels just outside the box extent, by $0.2$\,m at training and $0.5$\,m at inference, the larger margin compensating for the localization error of predicted boxes. A box that still contains no voxel after dilation is dropped from the training batch, where it would contribute neither a positive nor a negative, and is given a zero embedding at inference, where every detection has to be assigned.

An aggregation operator produces a single vector from the voxel features in $\mathcal{V}_t^i$. We use mean pooling,
\begin{equation}
\bar{f}^t_i = \frac{1}{|\mathcal{V}_t^i|} \sum_{v \in \mathcal{V}_t^i} f_v,
\label{eq:pool}
\end{equation}
which has no learnable parameters and stays well-conditioned across the wide range of $|\mathcal{V}_t^i|$, without amplifying the outlier features common under partial occlusion. A projection network $g_\psi$ with parameters $\psi$ then maps the aggregated vector to the final embedding, which is $L_2$-normalized:
\begin{equation}
e^t_i = \frac{g_\psi(\bar{f}^t_i)}{\|g_\psi(\bar{f}^t_i)\|_2}.
\label{eq:embedding}
\end{equation}
Normalization places $e^t_i$ on the unit sphere and makes the later similarity computations invariant to magnitude. The vector $e^t_i$ is the instance embedding of pedestrian $i$ at frame $t$.

\subsection{Contrastive Objective and Online Association} \label{sec:contrastive}

\paragraph{Training objective.}
At training time the ground-truth identities $\hat\tau$ define which embeddings should match. A batch is built from several frames so that a given pedestrian appears more than once. For an anchor embedding $e^{t}_i$ with identity $\hat\tau_i$, let $\mathcal{P}(i)$ be the set of other embeddings in the batch that share its identity (the positives) and $\mathcal{N}(i)$ the set of embeddings with a different identity (the negatives). Positives and negatives come from any frame in the batch and negatives may belong to a different sequence; how the frames are sampled is the main axis of variation between the configurations of Section~\ref{sec:configs}. The contrastive loss for one anchor is the multi-positive InfoNCE objective~\cite{infonce},
\begin{equation}
\ell_i = -\frac{1}{|\mathcal{P}(i)|} \sum_{e^+ \in \mathcal{P}(i)} \log \frac{\exp(e^{t}_i \cdot e^+ / \kappa)}{\sum_{e' \in \mathcal{P}(i) \cup \mathcal{N}(i)} \exp(e^{t}_i \cdot e' / \kappa)},
\label{eq:infonce}
\end{equation}
where $\kappa > 0$ is a temperature. Since all embeddings are $L_2$-normalized, each dot product equals cosine similarity and lies in $[-1, 1]$. With $|\mathcal{P}(i)| = 1$ this reduces to standard single-positive InfoNCE. The tracking loss averages \eqref{eq:infonce} over all anchors in the batch $\mathcal{A}$,
\begin{equation}
\mathcal{L}_{\text{track}} = \frac{1}{|\mathcal{A}|} \sum_{i \in \mathcal{A}} \ell_i.
\label{eq:track_loss}
\end{equation}

\paragraph{Online association at inference.}
At inference, identities are unavailable. Each active track is associated with a memory of its recent embeddings, $\mathcal{M}_\tau = \{\, e^{t'}_\tau : t' \in \mathcal{O}_\tau,\ t-H \le t' < t \,\}$, the embeddings from those frames within the horizon in which the track was observed, so a short gap in observation does not empty the memory. Given detections at frame $t$ with embeddings $\{e^t_i\}$, the similarity between detection $i$ and track $\tau$ is the maximum cosine similarity over that memory,
\begin{equation}
\text{sim}(i, \tau) = \max_{e' \in \mathcal{M}_\tau} e^t_i \cdot e',
\label{eq:sim}
\end{equation}
which defines a similarity matrix $S$ with entries $S_{i\tau} = \operatorname{sim}(i,\tau)$. Assignment solves a linear sum assignment problem with the Hungarian algorithm,
\begin{equation}
\pi^* = \arg\max_\pi \sum_i S_{i, \pi(i)} \quad \text{subject to} \quad S_{i, \pi(i)} \geq s_{\min},
\label{eq:hungarian}
\end{equation}
where $\pi$ ranges over partial matchings, injective maps that may leave detections and tracks unassigned, and $s_{\min}$ rejects weak matches. Unmatched detections initialize new tracks. There is no separate track-age parameter; the memory horizon $H$ plays that role. Embeddings older than $H$ frames are discarded, and a track whose memory has emptied can no longer be matched and leaves the active set. The pipeline uses no Kalman filter and no motion model; identity is carried across frames only by the learned embeddings, which isolates the effect of representation quality on tracking.

\subsection{Joint Fine-tuning of the Detector Backbone} \label{sec:joint}

With a frozen backbone, only the projection network $g_\psi$ adapts, while the features it reads stay fixed to what detection training produced. Letting the backbone adapt should help, since detection features separate pedestrians from background whereas tracking must separate one pedestrian from another, and our experiments (Section~\ref{sec:experiments}) confirm that this gives the largest single improvement.

Sharing the backbone between two objectives creates the standard multi-task difficulty. We optimize the fixed-weight joint loss
\begin{equation}
\mathcal{L}_{\text{joint}} = \mathcal{L}_{\text{det}} + \lambda\, \mathcal{L}_{\text{track}},
\label{eq:joint}
\end{equation}
with $\mathcal{L}_{\text{det}}$ the detector's own loss, $\mathcal{L}_{\text{track}}$ the contrastive loss of \eqref{eq:track_loss}, and $\lambda = 0.1$ held constant throughout training. Both the backbone parameters $\phi$ and the head parameters $\psi$ are updated from this objective.

The two objectives do not always agree. Detection favors features that are invariant across similar pedestrians, while tracking must separate those same pedestrians by identity, and the two gradients carry a negative cosine similarity on a sizable fraction of steps (Section~\ref{sec:exp_dynamics}). We also tried arbitrating this conflict explicitly, by projecting the tracking gradient onto the plane orthogonal to the detection gradient and by rescaling the two terms from their gradient norms. Neither changed results beyond run-to-run variation, so we report the plain fixed-weight objective as the method. The improvement therefore survives the conflict rather than depending on how it is arbitrated, which does not rule out that a better rule exists.

\subsection{Implementation Details} \label{sec:configs}

\paragraph{Backbones.}
We instantiate the feature extractor $\mathcal{F}_\phi$ with two LiDAR 3D pedestrian detectors. The primary backbone is DCCLA~\cite{dccla}, a sparse U-Net with dense cross-connections and linear attention, chosen for its state-of-the-art pedestrian detection on this benchmark. The secondary backbone is RPEA~\cite{rpea}, a residual-path detector with channel attention, included to check whether the approach depends on the internals of one architecture. Both expose the same interface to the tracker, a voxel-level feature map read before the detection head, and both are retrained from scratch in the tracker's codebase following their original protocols~\cite{dccla, rpea}. The projection network $g_\psi$ is a small MLP producing $d = 128$ dimensional embeddings.

\paragraph{Configurations.}
Five training configurations vary the frozen or unfrozen status of the backbone and the way frames are sampled to form positives; they differ only in which embeddings enter the positive set $\mathcal{P}(i)$ of \eqref{eq:infonce}, and negatives are always the other pedestrians in the batch. Configuration~A draws two augmented views of the same scan (rotation $\pm 10^\circ$, scale jitter $0.95$ to $1.05$, horizontal flip with probability $0.5$), which tests whether identity signal is present in detector features without temporal supervision. Configuration~B replaces them with two real frames of the same pedestrian at a gap of $1$ to $10$ frames. Configuration~C samples $K = 4$ frames of one sequence per batch with gaps up to $20$, so every pedestrian present in at least two of them contributes all its appearances to $\mathcal{P}(i)$. Configuration~D runs A, then B, then C as a warm-start curriculum, and Configuration~E keeps the $K = 4$ sampling of~C but unfreezes the backbone under \eqref{eq:joint}.

\paragraph{Training and inference settings.}
All experiments use the JRDB 3D tracking benchmark~\cite{jrdb}, described in Section~\ref{sec:experiments}. Frozen configurations train the head only, with batch size 64 for 160 epochs at a peak learning rate of $2\times10^{-3}$, starting from the detector checkpoint. Joint fine-tuning uses batch size 8 for 40 epochs at $2\times10^{-4}$, starting from the Configuration~C checkpoint; the smaller batch bounds the memory cost of backpropagating through the backbone and the lower rate keeps it from drifting away from the detection function. All runs use AdamW~\cite{adamw} with the one-cycle schedule~\cite{onecycle}, temperature $\kappa = 0.05$, embedding dimension $d = 128$, and voxel size $(0.05,0.05,0.10)$\,m~\cite{dccla,rpea}. Inference settings are fixed across configurations and tuned once on validation: memory horizon $H = 1000$ frames and minimum similarity $s_{\min} = 0.7$. Training uses PyTorch~\cite{pytorch} and TorchSparse~\cite{torchsparse} on a single 80\,GB GPU.

Checkpoints are selected on a proxy metric rather than on the contrastive loss. The proxy is the top-1 nearest-neighbor identity accuracy on a validation split: for each anchor embedding, the metric checks whether its nearest other embedding shares the same ground-truth identity. Across configurations this proxy correlates with the downstream HOTA and AssA, which is why it is used for selection; the contrastive loss, by contrast, keeps decreasing after downstream quality has stopped improving.
\section{Experiments and Results} \label{sec:experiments}


This section evaluates the configurations of Section~\ref{sec:configs} on the JRDB benchmark, first on the held-out test set and then through ablations on the validation set, and closes with a cross-dataset evaluation on SiT.

\subsection{Experimental Setup} \label{sec:exp_setup}

\paragraph{Dataset.}
JRDB~\cite{jrdb} is a 3D tracking benchmark recorded from a mobile robot on a university campus, which matches the social-navigation setting this work targets. Of its 54 sequences, 27 are annotated; we follow the official split of 20 training and 7 validation sequences, and submit to the held-out test set of 27 sequences whose annotations are private. Generalization beyond JRDB is assessed separately on SiT~\cite{sit}, a second robot-mounted pedestrian dataset, in Section~\ref{sec:exp_sit}.

\paragraph{Submission protocol.}
Test-set metrics are computed by the JRDB evaluation server, which limits each account to a small number of submissions per month. The validation set therefore carries the ablation study: configurations A through E are compared in one local run, while only the baseline and Configuration~E go to the test server.

\paragraph{Metrics.}
Four metrics are reported throughout, with standard definitions cited rather than restated. HOTA~\cite{hota} is the geometric mean of a detection accuracy DetA and an association accuracy AssA, so it cannot be raised by detection alone. MOTA~\cite{clearmot} aggregates per-frame missed boxes, false positives, and identity switches, and is dominated by detection errors. IDF1~\cite{idf1} measures long-range identity consistency through a global one-to-one matching between predicted and ground-truth tracks. OSPA$^{(2)}$~\cite{ospa} compares the track sets as whole trajectories and splits into localization $\text{OSPA}_{\text{LOC}}$ and cardinality $\text{OSPA}_{\text{CARD}}$ terms; lower is better. HOTA and IDF1 reward association, MOTA and $\text{OSPA}_{\text{CARD}}$ reward detection and set cardinality, so a method that changes association moves the first group more.

\paragraph{Detector reproduction.}
Both backbones are retrained from scratch in the tracker's codebase and reproduce the published detection AP within 0.3 percentage points at IoU thresholds $0.3$, $0.5$ and $0.7$, so association gains cannot be attributed to a stronger or weaker detector.

\subsection{Main Results on the JRDB Test Set} \label{sec:exp_main}

Table~\ref{tab:test} reports the test-set metrics for the three submitted models: the frozen-backbone baseline (Configuration~A) and the two joint fine-tuning variants of Configuration~E. The dominant pattern is a wide gap between A and E on every association-oriented metric. AssA rises from 20.07 to 32.41, a 61.5\% relative increase; AssPr more than doubles, from 32.96 to 70.06; IDF1 grows by 10.0 points; and IDSW falls from 54{,}642 to 21{,}757, a 60.2\% reduction. These changes come at almost no cost to detection: DetA decreases by 1.05 points and DetPr stays within 0.2 points.

The localization-trained backbone of Configuration~A produces voxel features good enough for boxes (DetA 47.10) but poor for identity (AssA 20.07); after joint fine-tuning it reaches a representation that supports both, which improves association by a wide margin while leaving detection close to intact.

\begin{table}[t]
\centering
\caption{JRDB 3D tracking test set. Configuration~A is the frozen-backbone baseline; Configuration~E uses joint backbone fine-tuning. Best per row in bold. Arrows give the direction of improvement. The transition from A to E adds 7.9 HOTA points and reduces identity switches by about 60\%, at a small detection cost. The OSPA$^{(2)}$ row is the one exception, analyzed in Section~\ref{sec:exp_ospa}.}
\label{tab:test}
\begin{tabular}{@{}lcc@{}}
\toprule
Metric & Config A & Config E \\
\midrule
HOTA\,$\uparrow$ & 30.70 & \textbf{38.58} \\
DetA\,$\uparrow$ & \textbf{47.10} & 46.04 \\
AssA\,$\uparrow$ & 20.07 & \textbf{32.41} \\
AssPr\,$\uparrow$ & 32.96 & \textbf{70.06} \\
MOTA\,$\uparrow$ & 40.24 & \textbf{42.55} \\
IDF1\,$\uparrow$ & 27.99 & \textbf{37.99} \\
IDSW\,$\downarrow$ & 54{,}642 & \textbf{21{,}757} \\
\midrule
OSPA$^{(2)}$\,$\downarrow$ & \textbf{0.764} & 0.861 \\
\quad OSPA$_{\text{LOC}}$\,$\downarrow$ & 0.477 & \textbf{0.162} \\
\quad OSPA$_{\text{CARD}}$\,$\downarrow$ & \textbf{0.287} & 0.699 \\
\midrule
Detections & 623{,}286 & 610{,}315 \\
Track IDs & 1{,}435 & 8{,}921 \\
\bottomrule
\end{tabular}
\end{table}

Table~\ref{tab:leaderboard} compares Configuration~E against the published methods on the JRDB 3D tracking leaderboard, whose values are taken from the public listing. CLIFT places first on HOTA (38.58 against 25.54, a margin of 13.04 points) and on IDF1 (37.99 against 28.69, a margin of 9.30), and second on MOTA (42.55 against 43.37) and OSPA$^{(2)}$ (0.861 against 0.783). The metrics that reward association move much more than those that primarily reward detection or set cardinality.

ELPTNet is the cleanest comparison: a tracking-by-detection pipeline on the same DCCLA detector, with handcrafted geometric, appearance and motion features in place of learned embeddings. The 13-point HOTA gap, with the backbone fixed, isolates the contribution of the learned representation.

\begin{table}[t]
\centering
\caption{JRDB 3D tracking leaderboard. Best in bold, second best underlined. Baseline values are taken from the public leaderboard. $^{\dagger}$~Entry on the public JRDB 3D tracking leaderboard~\cite{jrdb} without an associated publication.}
\label{tab:leaderboard}
\begin{tabular}{@{}lcccc@{}}
\toprule
Method & HOTA\,$\uparrow$ & MOTA\,$\uparrow$ & IDF1\,$\uparrow$ & OSPA$^{(2)}$\,$\downarrow$ \\
\midrule
DRFDFF$^{\dagger}$          & 24.78 & \textbf{43.37} & 27.47 & 0.903 \\
ELPTNet~\cite{elptnet}      & \underline{25.54} & 32.65 & \underline{28.69} & \textbf{0.783} \\
S2MAT~\cite{s2mat}          & 18.93 & 24.21 & 21.34 & 0.975 \\
SS3D-MOT~\cite{ss3dmot}     & 15.80 & 22.96 & 14.48 & 0.982 \\
PC-DAN~\cite{pcdan}         & 15.57 & 22.57 & 13.89 & 0.976 \\
AB3DMOT~\cite{ab3dmot}      & 11.81 & 19.34 & 10.67 & 0.971 \\
\midrule
CLIFT (Config~E), ours      & \textbf{38.58} & \underline{42.55} & \textbf{37.99} & \underline{0.861} \\
\bottomrule
\end{tabular}
\end{table}

\subsection{Ablation: Training Configurations A--E} \label{sec:exp_ablate}

Table~\ref{tab:ablate} reports validation metrics for all five configurations on the DCCLA backbone.

The transition from B to C is the largest single jump in the table: HOTA grows from 27.03 to 32.91, AssA from 20.83 to 30.22, and IDSW collapses from 63{,}678 to 23{,}235, a 63.5\% reduction. Sampling four frames per batch and using every same-identity pair across them supplies more identity supervision than the pairwise sampling of Configuration~B, and the resulting embeddings survive longer-horizon variation.

The transition from A to B is by contrast almost flat (HOTA 26.94 to 27.03): temporal pairs at gaps of 1 to 10 frames add little beyond augmentation, since short-horizon pedestrian motion resembles a small synthetic perturbation. Configuration~D, the A$\to$B$\to$C curriculum, lands within run-to-run noise of C (HOTA 33.11 against 32.91), so Configuration~E starts from the C checkpoint instead. The step from C to E adds 1.5 to 2 HOTA points by unfreezing the backbone.

\begin{table}[t]
\centering
\caption{Validation-set ablation over training configurations on the DCCLA backbone. Best per column in bold. A: same-frame augmentation, frozen. B: temporal pairs. C: $K{=}4$ multi-frame. D: A$\to$B$\to$C curriculum. E: joint fine-tune. The B$\to$C step is the largest jump; E adds a further gain on top of C; DetA is stable throughout.}
\label{tab:ablate}
\begin{tabular}{@{}lcccccc@{}}
\toprule
Config & HOTA\,$\uparrow$ & DetA\,$\uparrow$ & AssA\,$\uparrow$ & MOTA\,$\uparrow$ & IDF1\,$\uparrow$ & IDSW\,$\downarrow$ \\
\midrule
A & 26.94 & 37.07 & 20.82 & 56.88 & 27.89 & 84{,}959 \\
B & 27.03 & \textbf{37.40} & 20.83 & 59.35 & 27.54 & 63{,}678 \\
C & 32.91 & 37.26 & 30.22 & 64.03 & 37.07 & 23{,}235 \\
D & 33.11 & 37.31 & 30.41 & 64.18 & 37.21 & 22{,}890 \\
E & \textbf{34.43} & 37.25 & \textbf{32.99} & \textbf{64.27} & \textbf{39.23} & \textbf{21{,}147} \\
\bottomrule
\end{tabular}
\end{table}

\subsection{The OSPA Cardinality Trade-off} \label{sec:exp_ospa}

OSPA$^{(2)}$ is the one metric in Table~\ref{tab:test} on which Configuration~A outperforms Configuration~E, 0.764 against 0.861. The decomposition explains it: localization improves threefold, 0.477 to 0.162, while cardinality degrades by a factor of 2.4, 0.287 to 0.699, and the cardinality penalty dominates. The apparent cause, a sixfold rise in track identities from 1{,}435 to 8{,}921, is an artifact: the two test submissions used different association thresholds, so the counts are not comparable. Measured within a single ablation at a fixed threshold, Configuration~E fragments considerably less than Configuration~A rather than more. Mean track length rises from 17.8 to 113.1 frames, and the share of single-frame tracks falls from 49.2\% to 20.1\%.

The operating point that produces these counts was never tuned. Holding the trained model fixed and lowering the association threshold from the submitted $0.7$ to $0.65$ cuts the number of identities by 21\%, from 5{,}937 to 4{,}663, and raises IDF1 from 39.23 to 40.01, at a cost of 0.44 AssA points and 635 additional identity switches. Set cardinality is therefore governed by an inference parameter rather than by a limitation of the learned representation.

The switches that remain are not spread evenly. Matching predictions to ground truth at 0.5\,m, 42\% fall in the first or last 20\% of a trajectory, where people enter and leave the field of view, while 58\% occur mid-trajectory, where a lower threshold merges fragments but does not repair the break. Long-horizon re-identification after a full occlusion, which the memory of Section~\ref{sec:contrastive} does not attempt, is the remaining gap and the natural next step.

\subsection{Architecture Transferability} \label{sec:exp_rpea}

Repeating Configuration~A with RPEA~\cite{rpea} in place of DCCLA on the validation set tests whether the contrastive pipeline depends on the internals of one detector. The two backbones agree within 0.3 points on the metrics the head controls: HOTA 26.94 against 26.99, DetA 37.07 against 37.05, AssA 20.82 against 20.57, all below the noise floor of repeated runs. This is the expected outcome if the pipeline exploits detector-level voxel features that both architectures share rather than properties specific to one design. MOTA and IDSW differ more (56.88 against 60.05, and 84{,}959 against 53{,}716), but this follows the distinct detection statistics of the two backbones: RPEA produces fewer total detections here, which mechanically lowers the identity-switch count.

Running the full joint fine-tuning of Configuration~E on RPEA reproduces the pattern observed on DCCLA. Against those frozen Configuration~A numbers, HOTA rises from 26.99 to 30.45, AssA from 20.57 to 26.21, and identity switches fall from 53{,}716 to 31{,}672, a 41\% reduction, while DetA gains 0.41 points. The improvement is concentrated in association, as on DCCLA, at a smaller margin. Detection pays slightly more on the secondary backbone: at the epoch on which validation was logged, RPEA loses 0.87, 1.23, and 0.92 AP points at IoU thresholds 0.3, 0.5, and 0.7, against about 0.5 for DCCLA.

\subsection{Cross-dataset Evaluation on SiT} \label{sec:exp_sit}

We evaluate on SiT~\cite{sit}, chosen over the vehicle-mounted driving benchmarks because it is robot-mounted, pedestrian-only and recorded with the same sensor configuration as JRDB (two VLP-16 units), over 14 sequences, 2{,}800 frames, 50{,}294 annotated boxes and 442 tracks. Neither model is fine-tuned on SiT and both consume identical detections from our own detector (recall 0.5026), so the comparison isolates association. No published SiT results exist for the baseline, so we ran ELPTNet~\cite{elptnet} in house, which is what makes the identical-detection protocol possible.

CLIFT reaches HOTA 22.04, AssA 24.44 and IDF1 29.15, against 21.72, 23.75 and 28.68 for ELPTNet, whose association is driven by a Kalman motion model. The ordering is preserved, but the margin is far smaller than the 13-point HOTA gap on JRDB, and the reason is measurable: SiT clips are capped at 200 frames, the median track lasts 108 frames against 344 on JRDB, and crowd density is roughly half. Long-horizon association, what the embeddings exist for, is barely exercised and a motion model nearly suffices. The result is therefore best read as evidence for dense robot-centric scenes, the setting the method targets, rather than as a claim about 3D MOT in general.

\subsection{What Joint Fine-tuning Changes in the Features} \label{sec:exp_probe}

The evidence so far is outcome-based. To locate where identity information appears, we probe the features on the seven held-out validation sequences, over 213{,}160 anchors at a gap of one frame and 202{,}764 at ten, measuring the cosine similarity to the same identity $k$ frames later (intra) and the fraction whose nearest neighbour carries the correct identity (top-1), before and after the tracking head.

\begin{table}[t]
\centering
\caption{Feature probe on the held-out validation sequences. Cells give detector features / embedding. Joint fine-tuning improves both, and the margin grows with the temporal gap.}
\label{tab:probe}
\begin{tabular}{@{}lcccc@{}}
\toprule
 & \multicolumn{2}{c}{intra\,$\uparrow$} & \multicolumn{2}{c}{top-1\,$\uparrow$} \\
\cmidrule(lr){2-3}\cmidrule(lr){4-5}
Backbone & $t{+}1$ & $t{+}10$ & $t{+}1$ & $t{+}10$ \\
\midrule
frozen & 0.936 / 0.872 & 0.836 / 0.749 & 0.687 / 0.851 & 0.419 / 0.594 \\
joint  & 0.954 / 0.935 & 0.875 / 0.842 & 0.773 / 0.927 & 0.496 / 0.759 \\
\bottomrule
\end{tabular}
\end{table}

Joint fine-tuning changes the detector's own features rather than only the projection head: read before the tracking head, top-1 at one frame rises from 0.687 to 0.773. The embedding's contribution grows with the gap, adding 7.6, 14.4 and 16.5 top-1 points at one, five and ten frames, direct evidence for the long-horizon stability previously argued only through HOTA. The mechanism is intra-identity consolidation, not separation: same-identity similarity rises at every gap while the hardest-negative similarity is unchanged. This also separates the method from a 3D port of joint detection-and-ReID learning~\cite{jde,fairmot}, whose supervision is within-frame semantics.

\subsection{Training Dynamics} \label{sec:exp_dynamics}

Two diagnostics support the design of Section~\ref{sec:joint}. The cosine similarity between the detection and tracking gradients is negative on roughly 40 to 50\% of steps, confirming that the two objectives genuinely compete for the shared backbone. Detection AP, checked every five epochs, stays within 0.5 points of its pre-fine-tuning value at all three IoU thresholds on DCCLA, so the fixed-weight objective preserves detection while the backbone adapts.

\subsection{Runtime and Memory} \label{sec:exp_runtime}

On a single RTX 3090, with CUDA synchronisation around each stage, a frame costs 87.7\,ms (11.4\,FPS) at a peak of 196\,MB. Voxel pooling, similarity and Hungarian matching take 5.4\,ms, 6.2\% of the frame; the reused backbone dominates the rest. A Kalman baseline spends 3.4\,ms on association on the same hardware, so a learned representation costs a few milliseconds more and removes the need for a motion model. The implementation is plain Python with PyTorch and carries no inference-specific optimization, so these figures bound the cost rather than describe a tuned deployment.

 \section{Conclusions} \label{sec:conclusions}

CLIFT applies contrastive learning to the intermediate voxel-level features of a pre-trained LiDAR detector, with a frozen-backbone phase and a joint fine-tuning phase. On JRDB it ranks first on HOTA and IDF1 and second on MOTA and OSPA$^{(2)}$, with the gains concentrated on association-centric metrics, and the pattern survives a change of detector backbone under full joint fine-tuning. Probing the features shows the improvement is consolidation of identity inside the detector's own representation, not only in the projection head. The bottleneck for LiDAR-only pedestrian tracking is therefore the cross-frame stability of object-level representations rather than detection accuracy, and multimodal fusion is not required for competitive tracking in this setting. On SiT the ordering is preserved but the margin narrows, so we scope the result to dense robot-centric scenes rather than to general-purpose 3D MOT. Long-horizon re-identification after full occlusion remains open, as does deployment on a mobile robot to measure how identity-stable tracks support downstream prediction and planning.

\section{Acknowledgements}

The authors used Claude Opus 4.8~\cite{anthropic2026claudeopus} for language editing and for cross-checking related literature. All AI-assisted output was verified by the authors, who take full responsibility for the paper and the code.


%
%
\bibliographystyle{splncs04}
\bibliography{main}
\end{document}